\documentclass[preprint,11pt]{elsarticle}
\usepackage{lipsum}
\usepackage{hyphenat}
\usepackage{tikz}
\usetikzlibrary{shapes.geometric, arrows}

\tikzstyle{startstop} = [rectangle, rounded corners, 
minimum width=3cm, 
minimum height=1.3cm,
text centered, 
draw=black, 
fill=red!30]

\tikzstyle{io} = [trapezium, 
trapezium stretches=true, 
trapezium left angle=70, 
trapezium right angle=110, 
minimum width=3cm, 
minimum height=1cm, text centered, 
draw=black, fill=blue!30]

\tikzstyle{process} = [rectangle, 
minimum width=3cm, 
minimum height=1cm, 
text centered, 
text width=3cm, 
draw=black, 
fill=orange!30]

\tikzstyle{decision} = [diamond, 
minimum width=3cm, 
minimum height=1cm, 
text centered, 
draw=black, 
fill=green!30]
\tikzstyle{arrow} = [thick,->,>=stealth]
\usepackage{subfig}
\usepackage{float}
\usepackage{multirow}
\usepackage{multicol}
\usepackage{url}
\usepackage{textcomp}
\usepackage{graphics} 
\usepackage{inputenc}
\usepackage{babel}
\usepackage{amsthm}
\usepackage{amsmath,amsfonts,amssymb}
\usepackage{algorithm}
\usepackage{algpseudocode}
\usepackage{mathtools}
\usepackage{textcomp}
\DeclareMathOperator*{\argmin}{arg\,min}
\newcommand{\dst}{\displaystyle}
\usepackage{lipsum}
\numberwithin{equation}{section}
\numberwithin{figure}{section}
\numberwithin{table}{section}
\usepackage{enumerate}

\begin{document}
\begin{frontmatter}
\title{A Lightweight Randomized Nonlinear Dictionary Learning Method using Functional Link Network RVFL}

\author[inst1]{G.Madhuri}
\author[inst1]{Atul Negi}
\affiliation[inst1]{organization={OCR Lab, School of Computer and Information Sciences},
            addressline={University of Hyderabad}, 
            city={Hyderabad},
            postcode={ 500046},
            state={Telangana},
            country={India}.
            }
            
\begin{abstract}
	Kernel-based nonlinear dictionary learning methods operate in a feature space obtained by an implicit feature map, and they are not independent of computationally expensive operations like Singular Value Decomposition (SVD). This paper presents an SVD-free lightweight approach to learning a nonlinear dictionary using a randomized functional link called a Random Vector Functional Link (RVFL). 
The proposed RVFL-based nonlinear Dictionary Learning (RVFLDL) learns a dictionary as a sparse-to-dense feature map from nonlinear sparse coefficients to the dense input features. Sparse coefficients w.r.t an initial random dictionary are derived by assuming Horseshoe prior are used as inputs making it a lightweight network. Training the RVFL-based dictionary is free from SVD computation as RVFL generates weights from the input to the output layer analytically. Higher-order dependencies between the input sparse coefficients and the dictionary atoms are incorporated into the training process by nonlinearly transforming the sparse coefficients and adding them as enhanced features. Thus the method projects sparse coefficients to a higher dimensional space while inducing nonlinearities into the dictionary. For classification using RVFL-net, a classifier matrix is learned as a transform that maps nonlinear sparse coefficients to the labels. The empirical evidence of the method illustrated in image classification and reconstruction applications shows that RVFLDL is scalable and provides a solution better than those obtained using other nonlinear dictionary learning methods. 
\end{abstract}
\end{frontmatter}
\section{Introduction}\label{sec:rvflintro}
To our knowledge and understanding, existing nonlinear dictionary learning methods require Singular Value Decomposition (SVD) to be performed in each step making them computationally intensive. When large datasets of high dimensionality are trained, performing SVD in each iteration is memory-intensive. So, there is a need for a simpler solution  which attempts to induce nonlinearities present in the data distribution into the dictionary.  
The objective is to design a lightweight nonlinear dictionary method that is free of SVD. When the number of dimensions exceeds the number of observations, a lightweight network avoids overfitting. A randomized feedforward neural network, RVFL, with the higher-order effects incorporated through nonlinear transformations of sparse coefficients is proposed to learn a dictionary as a sparse-to-dense map. 
 The success of deep learning methods in computer vision applications is attributed to several layers of nonlinear transforms applied to the raw features and the convolution of those features along with error back-propagation. To apply the generalized delta rule (Backpropagation), input patterns are multiplied with a linear matrix of weights and transformed nonlinearly (e.g. using sigmoid) to generate input for the next (hidden) layer. These weights are updated by error backpropagation in each iteration. Such hierarchical feature representations from raw data capture the inherent higher-order dependencies of the data \cite{sparsefeaturelearningfordeepbeliefnetworks}. Instead, in an RVFL net, higher-order effects are incorporated through nonlinear transformations of inputs and adding them as enhanced neurons to the input layer. 
 \subsection{Highlights}
 \begin{itemize}
 	\item The Sparse representation problem is formulated as an SVD-free RVFL problem. 
 	 Using a single nonlinear transformation of sparse coefficients and concatenating them as enhanced features results in a light weight RVFL-net.
 	\item Dictionary learned is a sparse-to-dense feature transform, mapping enhanced sparse coefficients to the input features.
 	\item A nonlinear dictionary is derived as a solution to the RVFL net problem in a single step. Analytical solutions to the dictionary and the classifier.
  	\item RVFLDL is different from other nonlinear dictionary learning methods avoiding gradient-descent and error backpropagation to update the dictionary.
 \end{itemize}
 \subsection{Random Vector Functional Link (RVFL)}
 An RVFL net is a Single-Layer Feed-Forward Neural Network (SLFN) with random weights assigned to neural connections from the input to the hidden layer and the input to the output layer. The nonlinearly transformed features concatenated to the input features form enhanced patterns and are directly connected to the output layer. With direct input-output links, this functional link improves the accuracy while reducing the error \cite{Pao_func_link_demo1988}.
 
This paper proposes formulating the Dictionary Learning (DL) problem as a Random Vector Functional Link (RVFL) problem. The merits of RVFL include its simple architecture that allows the error function to be quadratic. Thus, Conjugate Gradient Descent (CGD) or matrix inversion methods can be used for fast minimization \cite{RVFL_1999}. RVFL is proved to be a universal approximator without overfitting problems \cite{RVFL_1999}.

\subsection{Framework of RVFLDL}
\begin{figure}
	\centering
	\includegraphics[height=7.8cm,width=11.8cm]{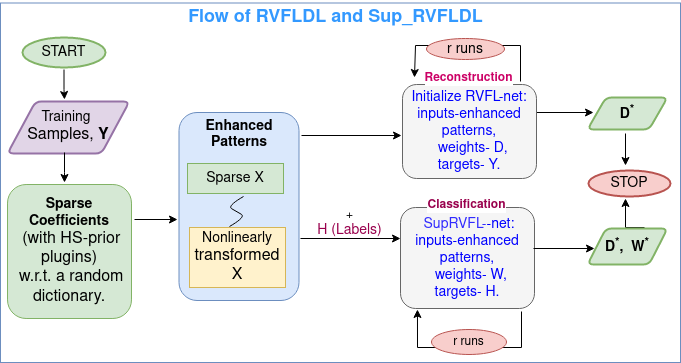}
	\caption{RVFLDL Framework}
	\label{fig:rvfl_framework}
\end{figure}

 The proposed formulation of using an RVFL net to derive a nonlinear dictionary from sparse coefficients is novel. The framework for the RVFL-based DL method to classify OCR image data is given in Fig. \ref{fig:rvfl_framework}.
RVFLDL has the following steps:
\begin{itemize}
    \item Sparsity-inducing Regularized Horseshoe prior \cite{Regularizedhorseshoeplugins} is assumed over the coefficient matrix to get sparse coefficients w.r.t. a random dictionary.
    \item A nonlinear function with random weights and biases is used to transform these coefficients nonlinearly. The sparse coefficients and their nonlinear transformed vectors are concatenated to form enhanced input patterns to the RVFL-net. Original training patterns are the targets to be approximated.  
    \item The dictionary is learned as the weight matrix mapping the enhanced nonlinear sparse coefficients directly to the dense target features. One such sparse-to-dense feature map from 3d-sparse to 2d-dense data has been used in Cross-modal learning for image annotation \cite{sparse_to_denseICCV2021}.
    \item An RVFL-net gives an analytical solution (using the Moore-Penrose pseudoinverse) to the dictionary (weight matrix between the enhanced input and output layers). If an iterative gradient-based method such as Conjugate Gradient (CG) is used, then the method converges in a finite number of iterations equal to the dimensionality of the weight matrix \cite{RVFLGeneralization}.
\end{itemize}
\subsection{Stability of the solution}
    However, using random weights and biases might lead to an unstable solution. So, a mean dictionary is obtained from $k-fold$ cross-validation where in each fold RVFLDL is run $r$ times, ($r$ is determined empirically) each time with random weights and biases to transform the coefficients nonlinearly. Thus, we have a set of $k$ classifiers and $k$ dictionaries. The mean of the dictionaries and the mean of the classifiers obtained from the $k-folds$ gives a stable solution.

In this paper, section \ref{sec:rvflPreliminaries} introduces the Sparse Representation (SR) problem, and Section \ref{sec:rvflrelatedwork} discusses related work. Section \ref{sec:RVFLDL} explains the objective function of the proposed RVFLDL method. Section \ref{sec:rvflcvgencecomplexity} presents the convergence and complexity analysis of RVFLDL and section \ref{sec:rvflexpresults} compares experimental results with other dictionary learning methods. Section \ref{sec:rvflDiscuss} and Section \ref{sec:rvflconclusion} discuss and conclude with the future scope of the work.
\section{Preliminaries}\label{sec:rvflPreliminaries}
The notation used in this article is presented in Table \ref{tab:RVFL_notation}.

\begin{table}[ht]
    \centering
    \caption{Notation}
    \label{tab:RVFL_notation}
    \begin{tabular}{|c|c|}
      \hline
       lower case alphabet, $\Vec{x}$& Vector \\
      \hline
      Upper case alphabet, $A$&Matrix  \\
      \hline
      $d$ & dimensionality of input, $Y$\\
      \hline
      $K$ & Size of initial dictionary, $D$\\
      \hline
      $N$ & No. of input patterns\\
      \hline
      $L$ & No. of hidden nodes\\
      \hline
      $\dst \|\Vec{a}\|_0=\#$non-zeros in $\Vec{a}$ & $l_0-$pseudonorm\\
      \hline
      $\dst \|\Vec{a}\|_p=\sum_{i}(|a_i|^p)^{\frac{1}{p}}$ & $l_p-$norm\\
      \hline
      $\dst \|A\|_F=\sqrt{\sum_{i}\sum_{j}a_{ij}^2}$ & Matrix Frobenious Norm\\
      \hline
    \end{tabular}
    
\end{table}

\subsection{Sparse Representation}\label{sec:sp_coding}
Let $Y\in \mathbb{R}^{d \times N}$ be the set of $N$ samples each of $d$ features (rows). The problem of Sparse Representation (SR) and  Dictionary Learning (DL) is to learn $X\in \mathbb{R}^{K\times N}$ and $D\in \mathbb{R}^{d\times K}$ to approximate $Y$. 
\begin{equation}\label{eq:rvfl_SR}
    <\hat{D},\hat{X}>= \argmin\limits_{D,X}{\|Y-DX\|_F^2+\lambda\|X\|_0}
\end{equation} where $X$ is sparse. 
This jointly non-convex problem \eqref{eq:rvfl_SR} is solved using the Block Coordinate Descent \cite{Block_coord_desc_Neurocomp2016} technique of updating one variable while fixing all other variables. The problem of updating $X$ w.r.t fixed $D$ is called Sparse Coding. 
\begin{equation}\label{eq:rvfl_sparsecoding}
    <X>=\argmin\limits_{X} \{\|Y-DX\|_F^2+\lambda G(X)\}, \forall j=1,2,\ldots,N
\end{equation} s.t. $\dst \|\Vec{d}_k\|_2^2=1, k=1,2,\ldots, K$.
When $G(.)=\|.\|_0$, greedy methods for sparse coding  \cite{GreedisgoodOMP,tropp2007signalOMP} exist to solve NP-Hard problem \eqref{eq:rvfl_sparsecoding}. When $G(.)=\|.\|_{1,1}=\sum_j{\sum_i(|G_{ij}|)}$, a matrix norm equivalent to the $l_1-norm$ for vectors, \eqref{eq:rvfl_sparsecoding} becomes \eqref{eq:rvfl_Lasso}.
\begin{equation}\label{eq:rvfl_Lasso}
    <X>=\argmin\limits_{X} \{\|Y-DX\|_F^2+\lambda \|X\|_{1,1}\}, \forall j=1,2,\ldots,N
\end{equation} s.t. $\dst \|\Vec{d}_k\|_2^2=1, k=1,2,\ldots, K$.

The LASSO problem \eqref{eq:rvfl_Lasso} is solved using methods like \cite{LARS,ISTA,FISTA2009}.
 In this work, we assume Horseshoe (HS) prior over the coefficients to obtain sparse coefficients of inputs w.r.t a random dictionary.

\subsubsection{Horse-Shoe prior \cite{carvalho2009first,carvalho2010horseshoesecond}}

\begin{figure}
    \centering
    \includegraphics[height=6cm,width=6.5cm]{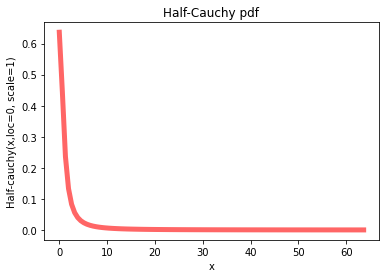}
    \caption{Half-Cauchy pdf with location 0 and peak width 1.}
    \label{fig:halfcauchy}
\end{figure}

Consider the problem where a given set of signals \[\Vec{y}_j=D\Vec{x}_j+\Vec{\epsilon}_j,\] $j=1,2,\ldots,N$. Additive Gaussian noise is assumed. Thus, \[\Vec{\epsilon}_j\sim \mathcal{N}((\Vec{y}_j-D\Vec{x}_j),\sigma^2 I),\] $j=1,2,\ldots,N$. The horseshoe prior is assumed on the coefficient matrix to get a sparse coefficient matrix. Here the coefficient vector $\Vec{x}$$_i$, corresponding to each dictionary atom $\Vec{d}$$_i$, is a scale mixture of Gaussians i.e. \[\dst (\Vec{x}_i| \lambda_i, \tau) \sim \mathcal{N}^+(0,\lambda_i^2\tau^2), i=1,2,\ldots,K\]. \[\lambda_i \sim \mathrm{C}^+(0,1)\], where $\mathrm{C}^+(0,1)$ is Half-Cauchy distribution. 
The Half-Cauchy distribution has a location parameter $\mu$. The peak’s width is dictated by a positive scale parameter $w$. 
Figure \ref{fig:halfcauchy} demonstrates the heavy-tailed Half-Cauchy distribution $\mathrm{C}^+(0,1)$. 
       Assuming Horseshoe prior over the coefficients, sparse coefficients are generated using \eqref{eq:HS-scoding}. The conditional posterior for the coefficients $X$, given the hyperparameters $\lambda_i, \tau, \sigma^2$ and data $\mathcal{D}=\{D, Y\}$ is given by \eqref{eq:HS-scoding}.
     \begin{equation}\label{eq:HS-scoding}
         p(X|\Lambda, \tau, \sigma^2, \mathcal{D})=\mathcal{N}(X|\Bar{X},\Sigma),
     \end{equation} where $\dst \Lambda = diag(\lambda_1^2, \lambda_2^2,\ldots,\lambda_K^2)$, $\tau$ is the global shrinkage parameter and $\lambda_i, i=1,2, \ldots, K$ are local shrinkage parameters. The following equations \eqref{eq:HS-1}, \eqref{eq:HS-2} and \eqref{eq:HS-3} from \cite{Regularizedhorseshoeplugins} are used to obtain the sparse coefficient matrix w.r.t. dictionary $D$.
     \begin{equation}\label{eq:HS-1}
      \Bar{X}=\tau^2 \Lambda(\tau^2 \Lambda + \sigma^2 (D^TD)^{-1})^{-1}\Hat{X}   
     \end{equation}
      \begin{equation}\label{eq:HS-2}
          \Sigma=(\frac{1}{\tau^2}\Lambda^{-1}+\frac{1}{\sigma^2}D^TD)^{-1}
      \end{equation}
      \begin{equation}\label{eq:HS-3}
          \Hat{X}= (D^TD+\eta I)^{-1}D^TY,
      \end{equation} where $\eta=0.01$.
Algorithm \ref{alg:sc-hs} is used to derive $X$ which serves as input to the RVFL net in the proposed RVFLDL algorithm. 
\begin{algorithm}
	\begin{algorithmic}[1]
		\caption{Sparse Coding with Horse-shoe prior \cite{Regularizedhorseshoeplugins}}\label{alg:sc-hs}
		\Procedure{RegularizedHoreseShoePlugins}{Normalized data $Y$, $\sigma^2, K, D\subset Y$ or $D=\{d_{ij}/d_{ij} \sim \mathcal{N}(0,1)\}, \lambda_j \sim \mathcal{C}^+(0,1), \tau $}
		
		\State{Output: $X$}
		\State{$\dst \Lambda \gets diag(\lambda_1^2, \lambda_2^2,\ldots,\lambda_K^2)$}
		\State{$\dst \Hat{X}\gets (D^TD+\eta I)^{-1}D^TY$}
		\State{$\dst \Bar{X}\gets \tau^2\Lambda(\tau^2\Lambda+\sigma^2((D^TD+\eta I)^{-1})^{-1}\Hat{X}$}
		\State{$\dst \Sigma \gets (\frac{1}{\tau^2}\Lambda^{-1} + \frac{1}{\sigma^2}D^TD)^{-1}$}
		\State{$\dst X \sim \mathcal{N}(\Bar{X},\Sigma)$ i.e. $X$ constitutes $N$ samples from this distribution.}
		
		\EndProcedure
	\end{algorithmic}
\end{algorithm}\vspace{-0.2cm}

 HS-prior takes care of global and local shrinkage by sampling them from the Half-Cauchy distribution. Thus, we avoid specifying the sparsity threshold.
\subsection{Dictionary Learning}
Dictionary learning problem updates $D$ by fixing $X$. 
\begin{equation}\label{dictionarylearning}
    \hat{D}=\argmin\limits_{D \in \mathcal{D}} \mathcal{F}(Y,D,X)+\mathcal{H}(W,D,X)+\lambda G(\Bar{x})\}, 
    \end{equation} s.t. $\dst \|\Vec{d}_k\|_2^2=1, k=1,2,\ldots, K$, where $\mathcal{D}$ is a set of matrices in $\mathbb{R}^{d\times K}$, $\mathcal{F}$, and $\mathcal{H}$ represent the data fidelity and discriminative parts of the objective.
Off-the-shelf linear analytic transforms as dictionaries were outperformed by learned dictionaries based on input signals for representation. 

 For nonlinearly separable data, the kernel method is used to transform K-SVD \cite{ksvd} and MOD \cite{MOD_Engan} into nonlinear dictionary learning methods for object recognition in \cite{NonlinDL_TIP2013}. A nonlinear dictionary is obtained as a product of the nonlinear transformation of the input matrix and the Moore-Penrose pseudo inverse of the coefficient matrix in \cite{PR2018NonlinDL_classification}. As stated earlier these methods involve iterative SVD operations.
To be able to capture the features of nonlinear data distributions using SR, several attempts have been made to either nonlinearly combine the dictionary atoms (\textit{Nonlinear Sparse Coding}) or learn a nonlinear dictionary (\textit{Nonlinear Dictionary Learning}) using kernel methods or deep neural networks with gradient descent and back-propagation.  

\section{Related Work}\label{sec:rvflrelatedwork}
The nonlinear dictionary learning problem is similar to the nonlinear Blind Source Separation (BSS) problem \eqref{eq:nonlin_BlindSourceSep} of extracting the unknown sources $x_i(t), i=1,2,\ldots, K$ nonlinearly transformed by the nonlinear function $g$ to generate the observations $y(t)$.
\begin{equation}\label{eq:nonlin_BlindSourceSep}
    y(t)=g(x(t)) +n(t),
\end{equation} where $n(t)$ is $d-$dimensional additive Gaussian noise at time $t$.
The BSS problem extracts the unknown sources $x_i(t),i=1,2,\ldots, K$ which generated the observations through a nonlinear mapping $g$. 
In the literature, neural networks and kernels have been used to introduce nonlinearities in the data into the dictionary learning process. 
We propose to use RVFL-net to derive a nonlinear dictionary. An \textit{RVFL net} is simple to use and has a rigorous mathematical justification (for all the functional links) given in \cite{RVFL_mathproof}. Learning and generalization characteristics of RVFL net are discussed in \cite{RVFLGeneralization}.
 In an RVFL net, the input layer is enhanced by adding the hidden layer to be part of the input i.e., the original pattern is concatenated to its nonlinear transformation to form an enhanced pattern. These enhanced nodes can be generated in large numbers initially and pruned later if they do not contribute much to the output or discrimination between classes \cite{RVFLGeneralization}. 
 To control the effect of randomized network parameters, the Sparse Pretrained RVFL (SP-RVFL) method trains a sparse autoencoder with $l_1-$regularization to learn the hidden layer parameters \cite{SPRVFL}. Hidden layer parameters in Random Weight Networks (RWNs) are sparsely coded using $l_1-$regularization in \cite{sparsecodedRWNs}. 
 The similarity of the RVFL solution to the weight matrix coincides with that of  Kernel Ridge Regression \cite{rvflKRRsimipnsuganthan}. 
Through an RVFL net, the nonlinear dictionary learning algorithm RVFLDL finds the mapping $D^T$ (the weight matrix in RVFL). $D$ is a mapping from the sparse coefficients $X$ and their nonlinear transformations as enhanced features to the original observations $Y$. 

\subsection{Kernel-based nonlinear dictionary learning} 
In \cite{NonlinDL_TIP2013}, the objective is to find $A$ such that \[\|\phi(Y)-\phi(Y)AX\|_F^2\] is minimized, i.e. the dictionary $A$ is to be learned in the higher dimensional feature space transformed using any Mercer kernel given by an implicit feature map $\phi(.)$. The authors demonstrated that USPS images are nonlinearly sparse w.r.t. the dictionary learned in the feature space transformed using a kernel. Even in Kernel Sparse Representation \cite{kernelSRTIP2013}, both input patterns and the dictionary are mapped to a higher dimensional feature space using Histogram Intersection Kernel, and sparse coding is performed using the Feature-sign search algorithm \cite{Efficientsparsecodingalgos}. 
In \cite{Nonlin_partiallabeldataPR2015}, a semi-supervised linear DL method is proposed and extended to nonlinear DL by applying a Mercer kernel on the input to transform them into a higher dimensional space. An Iterative Projection Method (IPM) with gradient descent at each step is used to solve the sparse coding problem, and classwise reconstruction errors are computed to determine the label of the test signal. However, the performance of kernel-based NDL methods is sensitive to the choice of the kernel which determines how the input features are mapped and measured in the feature space. Instead of mapped signals, pairwise signal similarities are accessible in the form of $K_{N\times N}$. As the number of observations $N$ increases, the memory required to store the kernel matrix explodes. 

\subsection{Neural network-based nonlinear dictionary learning} 
Neural network-based nonlinear dictionary learning methods apply linear dictionary learning methods (KSVD, MOD) to the nonlinear feature transformations. A feed-forward neural network is used to get hierarchical nonlinear projections of the data and the dictionary in \cite{PR2018NonlinDL_classification}. Here, an $(M-1)^{th}$ nonlinear transform of input i.e. the output of the middle ($M^{th}$) hidden layer is decomposed into the dictionary and coefficient matrices. The final ($2M^{th}$) output layer gives the reconstructed image of the input. However, the parameter optimization involves back-propagation, and gradient-descent. The alternative optimization methods increase the time complexity of the model. Also, there is no bound on the number $2M-2$ of hidden layers which might result in model overfit in the case of data scarcity.

\subsection{Kernel-regularized nonlinear dictionary learning } 
A Stacked AutoEncoder (SAE) is applied to get a low-dimensional feature map $f$ and a kernel is applied to get a high-dimensional feature map $\phi$. A nonlinear dictionary that minimizes \eqref{eq:kernelregularizedNDL} is learned with a joint sparsity constraint on the sparse codes obtained using dictionaries learned in both the feature spaces. Thus, the Kernel regularized NDL method \cite{Kernelregulaized_NDL_Trans_man_sys_cyber_2019}  optimizes the following objective:
\begin{equation}\label{eq:kernelregularizedNDL}
    \min\limits_{f,D, x_i,v_i} \sum_{i=1}^N {\|f(y_i)-Dx_i\|_2^2 + \lambda_R \mathcal{R}(f) + \lambda_K \|\phi(y_i)-\phi(Y)Av_i\|_2^2}
\end{equation}
s.t. $\|X_i\|_{row-0} \leq \tau$, $i=1,2,\ldots,N$.

 For each sample in the original space, two feature maps and the corresponding dictionaries and sparse codes are computed. Hence this method suffers from huge parametrization in SAE and implicit feature maps in kernel KSVD. 

\subsection{Broad learning framework}
In \cite{Broadlearning_addingenhancednodes_2018_trans_man_sys_cyber}, nonlinearity is added in the form of an enhanced nodes layer. Unlike direct input-output connections in RVFL-net, the broad learning framework has incremental algorithms employed to operate on mapped features to generate enhanced nodes.
The broad model is represented by
\[Y=[ Z_1,Z_2,\ldots,Z_d|H_1,H_2,\ldots,H_m]W^m\], where \[H_m=\psi(Z^dW_{h_{m}}+\beta_{h_{m}})\] denotes the $m^{th}$ group of enhancement nodes. \[Z_i=\phi(W_{e_{i}}^TX+\beta_{e_{i}}), i=1,2,\ldots,n.\] $W_{e_{i}}$ and $\beta_{e_{i}}$ are randomly generated and \[W^m=[Z^d|H^m]^+Y,\] $Z^d=Z_1,Z_2,\ldots,Z_d, H^m= H_1,H_2,\ldots,H_m$.\\
 In a broad learning framework, the number of hidden neurons to be added in each hidden layer to achieve a unique solution to its weight matrix is $N-rank(input)-1$ as given by \cite{Addinghiddenneuronsforfullrank_transonNN_1999}. Thus the number of learnable parameters is very high leading to an overfitting model when there are very few available observations.  When the number of observations is high and the rank of the input matrix is low, adding enhancement nodes in each layer becomes computationally expensive.\\

 \section{Proposed: \textbf{RVFLDL}}\label{sec:RVFLDL}
 This work is a novel application of RVFL to learn both the nonlinear dictionary and the classifier matrix for reconstruction and classification applications.
 The nonlinear dictionary learning problem in Random Vector Functional Link-based DL (RVFLDL) is to learn a dictionary \[\dst D^T:\mathbb{R}^{2K} \to \mathbb{R}^{d},\] with nonlinear components reflecting the nonlinearities in the data and to map the coefficients $\dst X\in \mathbb{R}^{2K\times N}$ to the actual observations $\dst Y\in \mathbb{R}^{d\times N}$. Sparse coefficients are nonlinearly transformed and an overcomplete dictionary is learned ($2K>d$) as a sparse-to-dense embedding. Sparse inputs to the RVFL lower the number of weight connections and hence the number of enhancement nodes.  

 To reduce dependence on hyperparameters in dictionary learning, we formulated the DL problem as an RVFL net problem, a nonparametric approach, that gives the dictionary a one-step solution without huge SVD computations present in other iterative dictionary learning methods. 
Consider the system of equations \[Y^T_{N\times d} \approx X^T_{N\times 2K}D^T_{2K\times d}\]
\[\iff Y_{d\times N}\approx D_{d\times 2K}X_{2K\times N}.\]

\subsection{Feature selection in RVFLDL} 
The proposed RVFLDL method where the regularization is imposed by the HS-prior on the coefficients achieves nonlinear dictionary learning objectives through the RVFL net without requiring any implicit feature maps or feature selection. Reducing the model complexity through sparse coefficients as inputs is equivalent to downsampling. \\
To compensate for the simple architecture of RVFL, and remove data redundancy, sparse coefficients of original features coded with a random dictionary are given as inputs. Sparse features extracted w.r.t a random dictionary avoid the handcrafted feature selection \cite{Lowrank_DL_unsupfeatselect_ES_2022}.

\begin{figure}
        \centering
        \includegraphics[height=9cm,width=12.8cm]{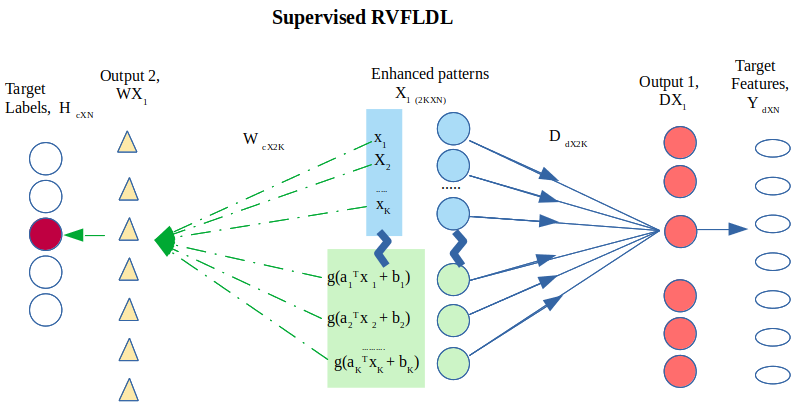}
        \caption{The proposed Supervised RVFLDL learns Dictionary and the classifier matrix simultaneously. Here, L=K.}
        \label{fig:sup_rvflfig}
    \end{figure}

 RVFLDL uses \eqref{eq:HS-1}, \eqref{eq:HS-2}, and \eqref{eq:HS-3}, to obtain sparse coefficient matrix w.r.t. a random dictionary $D$ as given in Algorithm \ref{alg:sc-hs}. 
 The global shrinkage parameter $\tau$ is usually set to 1 and the local shrinkage parameters are drawn from Half-Cauchy distribution with location 0 and scale 1. The RVFLDL approach does not require any implicit feature map (kernel) as the dictionary $D_1$ is learned to approximate sparse $X_1$ to low dimensional original dense features $Y \in \mathbb{R}^{d\times N}, K>d$. Regularization is imposed in the form of sparsity inducing Regularized HS prior, to the coefficients matrix $X$. 
The proposed RVFLDL operates on enhanced patterns obtained by concatenating nonlinear transformations of sparse coefficients to the sparse coefficients as shown in Fig \ref{fig:sup_rvflfig}.

To achieve a one-step solution to the dictionary, the sparse coefficient matrix $X$ has to be augmented to a full rank matrix by adding a bias node and $N-\alpha-1$ hidden nodes \cite{Addinghiddenneuronsforfullrank_transonNN_1999} where $\alpha$ is rank of $X$. 
Given a dataset \[Y \in \mathbb{R}^{d\times N}, \text{ and the coefficient matrix } X_1 \in \mathbb{R}^{(K+L) \times N},\] find $D_1 \in \mathbb{R}^{d \times (K+L)}$ such that \[\dst D_1X_1 \approx Y.\]  Here, the sigmoid function has been used to nonlinearly transform the coefficients to obtain an enhanced pattern i.e., \[\dst \Vec{x}_1=[\Vec{x} | \Vec{x}^1]\] where \[\Vec{x}^1_i=sigmoid(\Vec{w}_j^T\Vec{x_i}+ b_j).\]
The formulation of RVFLDL is given in 
\begin{equation}\label{eq:rvfl_formulate}
    \Vec{y}_i = \sum_{j=1}^{K}\Vec{d}_j x_{j} + \sum_{j=K+1}^{K+L}\Vec{d}_j\Vec{x}^1_j, i=1,2,,\ldots,N
\end{equation}
where $L$ corresponds to the number of hidden neurons in the RVFL net and $d_js$ are the weights to be learned. 
    Thus, the objective of RVFLDL is defined as 
     \begin{equation}\label{eq:rvflobjective}
        \mathbf{J}_{RVFLDL} = \min_{D_1}{\|Y-D_1X_1\|_F^2} + \mu_1 \|D_1\|_F^2,
     \end{equation} 
     where $X_1=[X, X']$ and $\mu_1$ is the regularization parameter penalizing the weight matrix $D_1$. 
     The solution to this convex optimization problem using Maximum Likelihood Estimation (MLE) is given by 
     \begin{equation}\label{eq:rvfl_MLE}
         D_1 = YX_1^T(X_1X_1^T+\mu_1 I)^{-1}
     \end{equation}
Sparsity constraint is omitted in \eqref{eq:rvflobjective} as $l_1-$norm regularization makes the function non-quadratic. Instead, the sparse coefficient matrix $X$ is learned in advance w.r.t. a random dictionary by assuming Horseshoe prior over the coefficients. Pre-learned sparse coefficient matrix embeds input information into the RVFL net giving it a better start.

\subsection{Supervised RVFLDL for Classification}
Let $\dst H \in \mathbb{R}^{c \times N}$ be the label matrix corresponding to the samples in $Y$.
Embed label information in the classifier matrix using RVFL. For the same matrix $\dst X_1=[X, X']$, use the label matrix $H$ as the output layer.
The formulation of obtaining the classifier matrix as the output weight matrix $W$ connecting the enhanced input sparse coefficients and the corresponding labels is given by 
\begin{equation}\label{eq:rvfl_classify_formulate}
    \Vec{h}_i = \sum_{j=1}^{K}\Vec{m}_j x_{j} + \sum_{j=K+1}^{K+L}\Vec{m}_j\Vec{x}^1_j, i=1,2,,\ldots,N.
\end{equation}

Thus, the classifier matrix is the result of 
    \begin{equation}\label{eq:suprvfldl}
        \mathbf{J}_{supRVFLDL} = \min_{W}{\|H-WX_1\|_F^2} + \mu_2\|W\|_F^2,
     \end{equation} 
 where $X_1=[X, X']$ and $\dst W=[\Vec{m}_j]_{j=1}^{K+L}$. The solution to this convex optimization problem using Maximum Likelihood Estimation (MLE) is given by 
     \begin{equation}\label{eq:rvfl_classifier}
         W = HX_1^T(X_1X_1^T+\mu_1 I)^{-1},
     \end{equation} where $\mu_2$ is the regularization parameter penalizing the weight matrix $W$.
 Using the Lagrange Multiplier Method, the objective function is 
      \begin{equation}\label{eq:supunsupRVFLDL}
      \begin{split}
         <\Hat{D_1},\hat{X_1},\hat{W}>&=\argmin_{D_1,W,X_1}\{\|Y-D_1X_1\|_F^2+\mu_3\|H-WX_1\|_F^2+\\
          &+\mu_1\|D_1\|_F^2+\mu_2\|W\|_F^2 \}
      \end{split}
     \end{equation}

\begin{algorithm}
	\caption{Proposed Supervised RVFLDL}\label{alg:suprvflDL}
	\begin{algorithmic}[1]
		\Procedure{SupRVFLDL}{Input: Normalized data $Y$, $\sigma^2, X$ from Algorithm \ref{alg:sc-hs}, $\mu_1$, $\mu_2$, $\#runs=r$,$\#folds = T$}
		\State{Output: $\dst D_1,W,label(query)$}
		\State{$\dst X`\gets g(\Vec{w}_i^Tx_i +$$ b_i), i=1,\ldots,K$ where $\dst \Vec{w}_i \sim \mathcal{N}(\Vec{0}$$,I)$$, b_i\sim \mathcal{N}(0,1)$.}
		\State{$\dst X_1 \gets [X,X`]$}
		\While{$t\leq T$}
		\While{$l \leq r$}
		\State{$D_l$ from \eqref{eq:rvfl_MLE}.}
		\State{$\dst X_1 \gets D_l^T(D_lD_l^T+\mu_1 I)^{-1}Y$ or from \eqref{eq:X1-MLE} for classification.}
		\State{$\dst W_l$ from \eqref{eq:rvfl_classifier}.}
		\EndWhile
		\State{$D_t \gets mean(D_l),l=1,\ldots,r$}
		\State{$W_t \gets mean(W_l),l=1,\ldots,r$}
		\EndWhile  
		\State{$D_1 \gets mean(D_t),t=1,\ldots,T$}
		\State{$W \gets mean(W_t),t=1,\ldots,T$}
		\State{Compute $X_1$ from \eqref{eq:X1-MLE} and $x_{query}$ from \eqref{eq:test-classify}.}
		\State{return label($query$)= SVC($X_1, x_{query}$).}
		
		\EndProcedure
	\end{algorithmic}
\end{algorithm}\vspace{-0.2cm}
Randomizing the weight vectors and bias results in an unstable solution. For a stable solution, steps 6 to 10 of Algorithm \ref{alg:suprvflDL} are repeated r times over a $T-$fold cross-validation, and a mean dictionary and classifier are obtained from the $T-$dictionaries and classifiers. The coefficient matrix $X_1$ is derived from the Maximum Likelihood Estimate in 
\begin{equation}\label{eq:X1-MLE}
\dst \frac{\partial J}{\partial X_1}=0 \implies X_1 = (D_1^TD_1+\mu_3W^TW+\eta I)^{-1}(D_1^TY+\mu_3W^TH)
\end{equation}

Support Vector Classifier is trained using the coefficients computed from \eqref{eq:X1-MLE} and a new test sample $\Vec{y}$$_q$ is classified using the coefficient vector $\Vec{x}$$_q$ computed from 
\begin{equation}\label{eq:test-classify}
x_q = (D_1^TD_1+\mu_3W^TW+\eta I)^{-1}D_1^Ty_q
\end{equation}

\section{Notes on Convergence and Complexity of RVFLDL}\label{sec:rvflcvgencecomplexity}
RVFL behaves like MLP when the function to be approximated is Lipschitz continuous \cite{RVFL_1999} i.e. the approximation error convergence is of $\dst O(\frac{1}{n})$. While other parameters are chosen randomly and are fixed, RVFL adapts only output weights, unlike MLP where all the parameters (hidden weights, output weights, and hidden thresholds ) are updated in the learning process. This avoids the local minima problem. The global minimum if exists is achieved using either matrix inversion or Conjugate Gradient Descent method. If the Hessian of the quadratic error function is positive definite, then Newton's method could also be used to achieve the global minima \cite{RVFL_1999}. For practical purposes, we have used the Kernel Ridge Regression \cite{rvflKRRsimipnsuganthan} form of solution where a small positive constant $\eta=0.01$ is added to all the diagonal entries of $X_1^TX_1$ to make it invertible. Thus the method converges for all the datasets. 

When compared to other dictionary learning methods, RVFLDL has no expensive SVD operation. However, the number of runs required to achieve better accuracies has to be decided empirically.
The number of weights to be learned in a single hidden layer Multi-Layer-Perceptron (MLP) with back-propagation is \[(i+1)*h+(h+1)*o\]. Here, $i,h,$ and $o$ indicate the number of neurons in the input, hidden, and output layers. With random activation weights and biases, RVFLDL has only \[(i+h)*o\] weights to be learned. Sparse coding using Regularized HS plugins involves $O(2K^3)$ multiplications. The complexity of RVFLDL involves learning the dictionary weights which involves multiplications of \[O((K+L)*d*N*r)\]. However, $X_1$ is sparse and the upper limit of nonzero components is $s$. Thus, the complexity becomes $O(2s*d*N*r)$. 
For Supervised RVFLDL, the computation of classifier matrix $W$ is added, which amounts to \[O(2K^3+(K+L)*(d+C)*N*r).\]
The numerical values in Table \ref{tab:RVFLDL_complexity} denote the worst-case values for the numbers considered here. 
\begin{table*}[ht]
	\centering
	\caption{Complexity analysis ($\#$multiplications) with the best complexities in column 2, worst-case complexities and their plug-in values in column 3 when $N=5*10^4, d=500, C=500, K=1000, s\leq30,r=10$, $N>K>d>s$.}
	\label{tab:RVFLDL_complexity}
	\begin{tabular}{|p{0.8in}|p{1in}|c|c}
		\hline
		\multirow{2}{*}{Module} & \multirow{2}{*}{Complexity} &  Worst over $r$ runs\\
		\cline{3-3}
		& & Plugin values\\
		
		\hline
		RVFLDL&$O(2s*d*N*r+2K^3)$ &$O((K+L)*d*N*r+2K^3)$\\
		\cline{3-3}
		& & $0.502\times 10^{12}$\\
		\hline
		Supervised RVFLDL&$O(2K^3+(2s)*Nr*(d+C))$&$O(2K^3+(K+L)*Nr*(d+C))$\\
		\cline{3-3}
		& & $0.1002\times10^{13}$\\
		\hline
	\end{tabular}
\end{table*}
These complexity values are less than the ones achieved by linear online DLSI $(0.625\times 10^{13})$ \cite{LRSDL} for $N=2000, K=1000, d=500, C=100, s=50$.
\section{\vspace*{-0.2cm}Experiments and Results}\label{sec:rvflexpresults}
\begin{figure*}
	\centering
    \subfloat[Sparse coefficient vector w.r.t random dictionary of size 200]{\includegraphics[height=6cm,width=6.1cm]{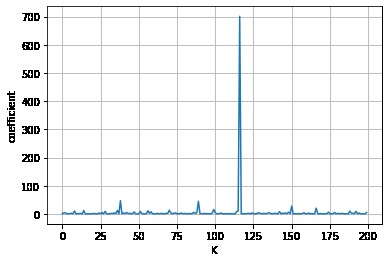}}
    \subfloat[Sparse coefficient vector w.r.t RVFLDL dictionary of size 400]{\includegraphics[height=6cm,width=6.1cm]{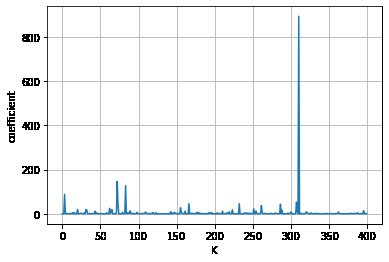}}
    \caption{Sparsity level is not altered using RVFLDL, but the  sparsity profile differs.}\label{fig:sparsecoefficients_comparison}
\end{figure*}

 Experimental setup: Memory 31.0 GiB, Intel® Core™ i7-10510U CPU @ 1.80GHz × 8, Ubuntu 20.04.6 LTS 64-bit.
The results are compared with other nonlinear dictionary learning methods in the literature, Kernel K-SVD, and Kernel MOD (KMOD) where the class label is decided based on the minimal reconstruction error.
The sparsity of coefficient vectors is achieved by assuming HS-prior. In Fig. \ref{fig:sparsecoefficients_comparison}, we observe that the sparsity of coefficients is not greatly modified using RVFLDL.  However, the sparsity profile (the set of dictionary atoms used) differs between using RVFLDL and without using RVFLDL.
Sparsity levels achieved using HS-prior over coefficients of different datasets, without and with RVFL dictionary are given in Table \ref{tab:RVFL_sparsity}.
\begin{table*}[ht]
    \centering
    \caption{RVFLDL does not much alter the level of sparsity. Dictionary size and the corresponding sparsity levels are given without and with RVFLDL dictionaries in column 3 and column 4.}
    \label{tab:RVFL_sparsity}
    \begin{tabular}{|c|c|c|c|}
        \hline
         Dataset &dim. & $K$, Sparsity &$(K+L)$, Sparsity  \\
         \hline
         USPS &256 &  200, 4&400, 7 \\
         ARDIS&784 & 450, 5&900, 6\\
         MNIST&784&450, 4&900, 6\\
         UHTelPCC& 1024&1200, 15& 2048,17\\
         EXt.YaleB &$54\times 48$& 900, 18 & 1800, 20\\
         \hline
    \end{tabular}
    
\end{table*}
The hyperparameters to be tuned while using RVFLDL are given in Table \ref{tab:RVFLparameters}. Global shrinkage, $\tau$ is set as 1.
         
\begin{table}[ht]
    \centering
     \caption{Parameters of RVFLDL}
    \label{tab:RVFLparameters}
    \begin{tabular}{|c|c|}
        \hline
         Parameter & Values  \\
         \hline
         Local shrinkage, $\lambda_i, i=1,2,\ldots,K$ & $\lambda_i \sim \mathcal{C}^+(0,1)$\\
         \hline
         Runs, r & Empirical tuning\\
         \hline
         Regularization parameters, $\mu_1,\mu_2,\mu_3$ & Tuned empirically within $[0,1]$\\
         \hline
    \end{tabular}
   
\end{table}
The global shrinkage parameter $\tau$ pulls all the variables towards zero while the local shrinkage parameters $\lambda_i, i=1,2,\ldots, K $ avoid the corresponding variables from being zero. However, a single value of $\tau$ may not work for all the datasets and is set empirically here.

\subsection{Classification using RVFLDL}
The classification performance of RVFLDL is tested using handwritten numeral OCR image datasets, MNIST, USPS, and ARDIS, which have intra-class variations. 
\subsubsection{MNIST (\cite{CirecsanMNISTICDAR2011})} MNIST is a benchmark dataset of hand-written numeral OCR images with intra-class variations. Each greyscale image is a $28\times 28$ image constituting $60000$ training and $10000$ testing sets.
\subsubsection{ARDIS \cite{kusetogullari2020ardis}}
\begin{figure*}[h]
\subfloat[Linear dictionary atoms of ARDIS images RVFLDL dictionary.]{\includegraphics[height=6cm,width=6cm]{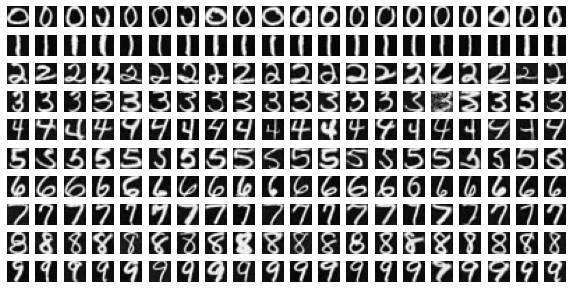}}
\subfloat[Non-linear dictionary atoms of ARDIS images RVFLDL dictionary.]{\includegraphics[height=6cm,width=6cm]{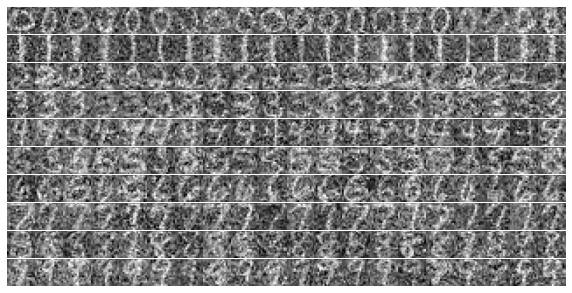}}
    \caption{Both linear and nonlinear atoms constitute RVFLDL dictionary}\label{fig:lin_nonlin_dictatoms}
\end{figure*}

ARDIS is a collection of Swedish church document OCR images of handwritten numerals. Each greyscale image is a $28\times 28$ image constituting $6600$ training and $1000$ testing sets. Compared to the dimensionality, the training data size is small. The learned atoms of the RVFLDL dictionary for ARDIS OCR images presented in Fig. \ref{fig:lin_nonlin_dictatoms} comprise those corresponding to the sparse coefficients and their nonlinear transformations.
 
\subsubsection{USPS \cite{USPS2020global}}
USPS constitutes greyscale images of size $16\times 16$ of handwritten numerals. 
It is empirically concluded that a degree 2 polynomial kernel-based SVC when $c=1$ performs better on OCR image datasets.
\subsubsection{UHTelPCC \cite{rakeshrtip2r}}
\begin{figure}[h]
    \centering
    \includegraphics[height=5cm,width=10cm]{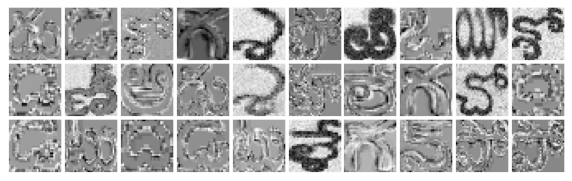}
    \caption{Dictionary atoms of the UHTelPCC dataset learned using Sup-RVFLDL.}
    \label{fig:UHTelPCC_RVFL_dictatoms}
\end{figure}

UHTelPCC (\textit{University of Hyderabad Telugu Printed Connected Component}) is a collection of $32\times 32$ OCR images of Telugu printed characters from 325 classes. Like other Dravidian scripts, Telugu has a complex orthography and several confusing classes. The training set has 50000 images and the test set has 10000 images. Figure \ref{fig:UHTelPCC_RVFL_dictatoms} presents sample atoms learned using the Supervised RVFLDL dictionary for the UHTelPCC dataset.
\subsubsection{Extended YaleB \cite{ExtYaleBPAMI2001}}
The face recognition ability of RVFLDL is tested using the Extended YaleB dataset of dimensionality $640 \times 480$ resized to $192\times 168$. The Extended Yale B face database contains 13102 face images collected from 28 people with different expressions on faces, occlusions, and light variations. Each subject has 460 images. 
\subsubsection{Cropped YaleB \cite{ExtYaleBPAMI2005}}
The Cropped Yale B face database contains 1939 face images collected from 38 people with different expressions on their faces, occlusions, and light variations. Each subject has around 50 images.

 Table \ref{tab:rvfldl_compare_NDL} compares the classification performance of RVFLDL with other nonlinear dictionary learning methods SNDL \cite{PR2018NonlinDL_classification}, NDL \cite{PR2018NonlinDL_classification}, Kernel KSVD \cite{NonlinDL_TIP2013}, KMOD \cite{NonlinDL_TIP2013}. SNDL and NDL use an encoder-decoder approach. The $M^{th}$ hidden layer in a $2M-$layer neural network is decomposed into dictionary and coefficient matrices. All the parameters of the model are updated using batch gradient-descent and back-propagation. Kernel KSVD and KMOD apply KSVD and MOD in the high-dimensional feature space obtained by an implicit Mercer kernel.
\begin{table}[ht]
         \centering
         \caption{RVFLDL classification performance is better than deep-neural network-based DL and Kernel-based nonlinear DL methods.}
         \label{tab:rvfldl_compare_NDL}
         \small
         \begin{tabular}{|p{0.5in}|p{0.6in}|p{0.5in}|p{0.5in}|p{0.5in}|p{0.5in}|p{0.5in}|}
         \hline
              Method&UHTelPCC&Ext.YaleB& CropYale&ARDIS& USPS & MNIST \\
              \hline
              SNDL&$98.1\pm.12$&$97.9\pm.69$&$95.39\pm 0.34$&$96.1\pm.54$&$94.7\pm.55$&$97.9\pm.92$\\
              \hline
              NDL&$92.3\pm.32$&$97.65\pm .49$&$95\pm 0.39$&$96.1\pm.44$& $95.1\pm.15$& $94.99\pm.52$\\
              \hline
              Kernel KSVD&$97.1\pm.54$&$98.45\pm.96$&$94\pm 0.59$&$95.1\pm.64$& $97.7\pm.66$&$97.9\pm.89$\\
              \hline
              KMOD&$97.1\pm.46$&$98.39\pm.79$&$93.4\pm 0.41$&$94.91\pm.44$&$94.2\pm.35$&$96.9\pm.86$\\
              \hline
              RVFLDL&$98.9\pm.3$&$98.69\pm.59$&$95.4\pm 0.33$&$96.91\pm.51$&$96.1\pm.35$& $97.99\pm.36$\\
              \hline
         \end{tabular}
         
     \end{table}

\begin{figure*}
\subfloat[]{\includegraphics[height=5cm,width=6cm]{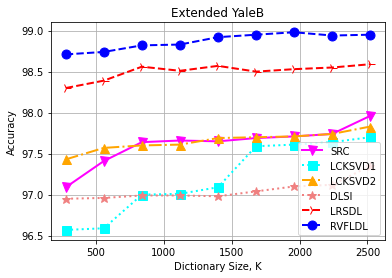}}
\subfloat[]{\includegraphics[height=5cm,width=6cm]{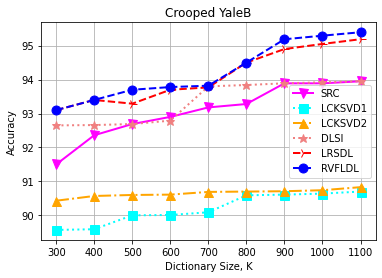}}\\
\subfloat[]{\includegraphics[height=5cm,width=6cm]{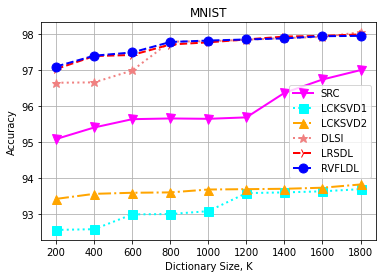}}
    \subfloat[]{\includegraphics[height=5cm,width=6cm]{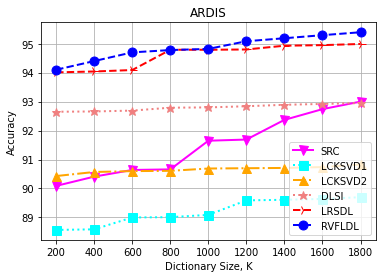}}\\
    \subfloat[]{\includegraphics[height=5cm,width=6cm]{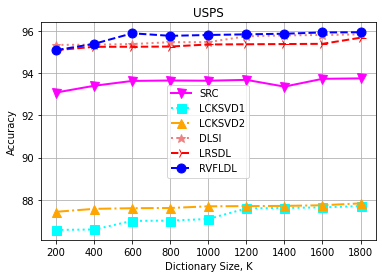}}
    \subfloat[]{\includegraphics[height=5cm,width=6cm]{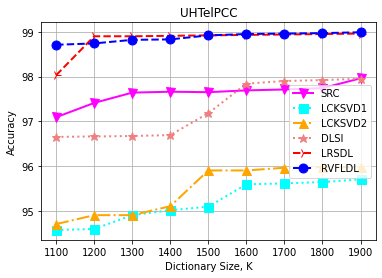}}
   \caption{RVFLDL performs better with few samples from each class when compared to other supervised linear dictionary learning methods}
    \label{fig:Acc_comparisongraph}
\end{figure*}

Figure \ref{fig:Acc_comparisongraph} shows that RVFLDL performs well with a relatively low number of dictionary atoms when compared to other linear dictionary learning methods DLSI \cite{DLSI}, SRC \cite{SRC}, LCKSVD1, LCKSVD2 \cite{LCKSVD}, LRSDL \cite{LRSDL}. 
\subsubsection{Parameter sensitivity analysis}
\begin{figure*}
\subfloat[]{\includegraphics[height=5cm,width=6cm]{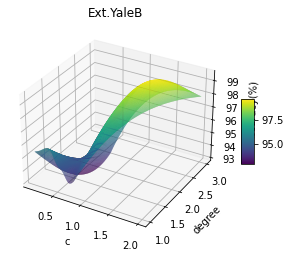}}
\subfloat[]{\includegraphics[height=5cm,width=6cm]{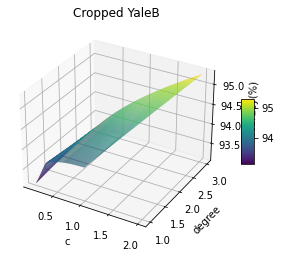}}\\
\subfloat[]{\includegraphics[height=5cm,width=6cm]{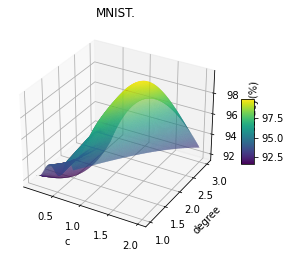}}
    \subfloat[]{\includegraphics[height=5cm,width=6cm]{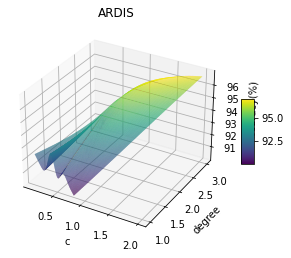}}\\
    \subfloat[]{\includegraphics[height=5cm,width=6cm]{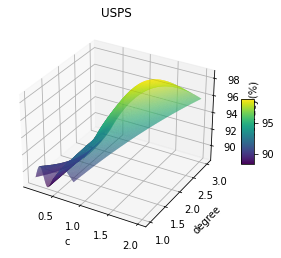}}
    \subfloat[]{\includegraphics[height=5cm,width=6cm]{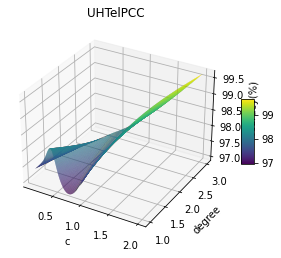}}
    \caption{Effect of c and degree of polynomial kernel of SVC on accuracy, using cubic interpolation. A higher degree polynomial kernel when c=1 gives better results.}\label{fig:effectofcanddegree_acc}
\end{figure*}

The dictionary size for each dataset varies with the number of samples per class used to train the dictionary. The sparsity threshold for each dataset is not specified, as the sparsity-inducing HS prior assumed over the coefficient matrix ensures their sparsity.
The effect of varying parameters on classification accuracy is studied. Fig. \ref{fig:effectofcanddegree_acc} presents the effect of parameters in SVC on the accuracy. A 3-fold cross-validation is performed with a polynomial kernel of degree 3 for classification where c is the regularization parameter of SVC, generally between [0.1,10). The regularization parameter is inversely proportional to the number of support vectors in SVC.

\begin{figure*}
\subfloat[]{\includegraphics[height=5cm,width=6cm]{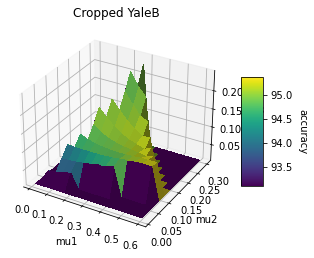}}
\subfloat[]{\includegraphics[height=5cm,width=6cm]{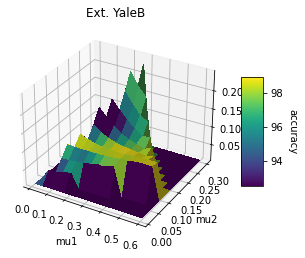}}\\
\subfloat[]{\includegraphics[height=5cm,width=6cm]{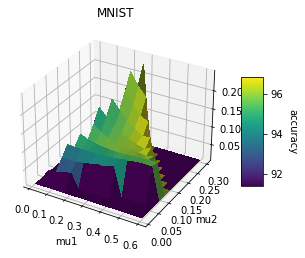}}
    \subfloat[]{\includegraphics[height=5cm,width=6cm]{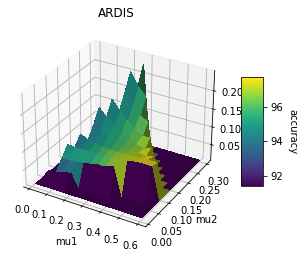}}\\
    \subfloat[]{\includegraphics[height=5cm,width=6cm]{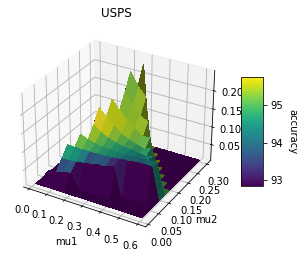}}
    \subfloat[]{\includegraphics[height=5cm,width=6cm]{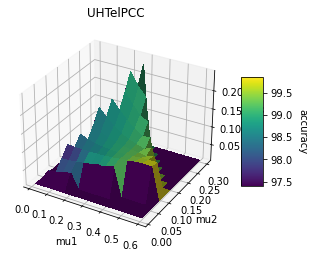}}
    \caption{Sup-RVFLDL: Effect of $\mu1,\mu2,\mu3$ on the classification accuracy, using cubic interpolation. $\mu_1=0.2,\mu_2=0.2,\mu_3=0.1$ gave better classification results. For the ARDIS dataset, $\mu_1=0.4,\mu_2=0.2,\mu_3=0.15.$}\label{fig:effectofmu_acc}
\end{figure*}

The parameters $\mu_1,\mu_2,\mu_3$ respectively penalize the dictionary weights, classifier weights, and the classifier term in the objective function \eqref{eq:supunsupRVFLDL}. Fig. \ref{fig:effectofmu_acc} presents the effect of the values of these parameters on the classification accuracy. For $\dst \mu_1$ in $\dst \{0,0.1,0.2,0.3,0.4,0.5,0.6\}$, $\dst \mu_2$ in $\dst \{0,0.05,0.1,0.15,0.2,0.25,0.3\}$, and $\dst \mu_3$ in $\dst \{0,0.05,0.1,0.15,0.2\}$, the parameter values $\mu_1=0.2,\mu_2=0.2,\mu_3=0.1$ gave better classification results. However, for the ARDIS dataset, $\mu_1=0.4,\mu_2=0.2,\mu_3=0.15$ gave better results.
\subsection{Image representation using RVFLDL}
An unsupervised RVFLDL dictionary is learned using \eqref{eq:rvflobjective} to reconstruct images. The sparse coefficients learned w.r.t. the unsupervised dictionary are used to reconstruct images using very few atoms of the overcomplete dictionary. The reconstruction results on the following datasets are compared with those of another  Kernel KSVD \cite{NonlinDL_TIP2013} method.
\subsubsection{COIL100 \cite{COIL100}} 
\begin{figure*}
\centering
    \subfloat[Reconstruction using RVFLDL, with SSIs close to 1.]{\includegraphics[height=5cm,width=9.3cm]{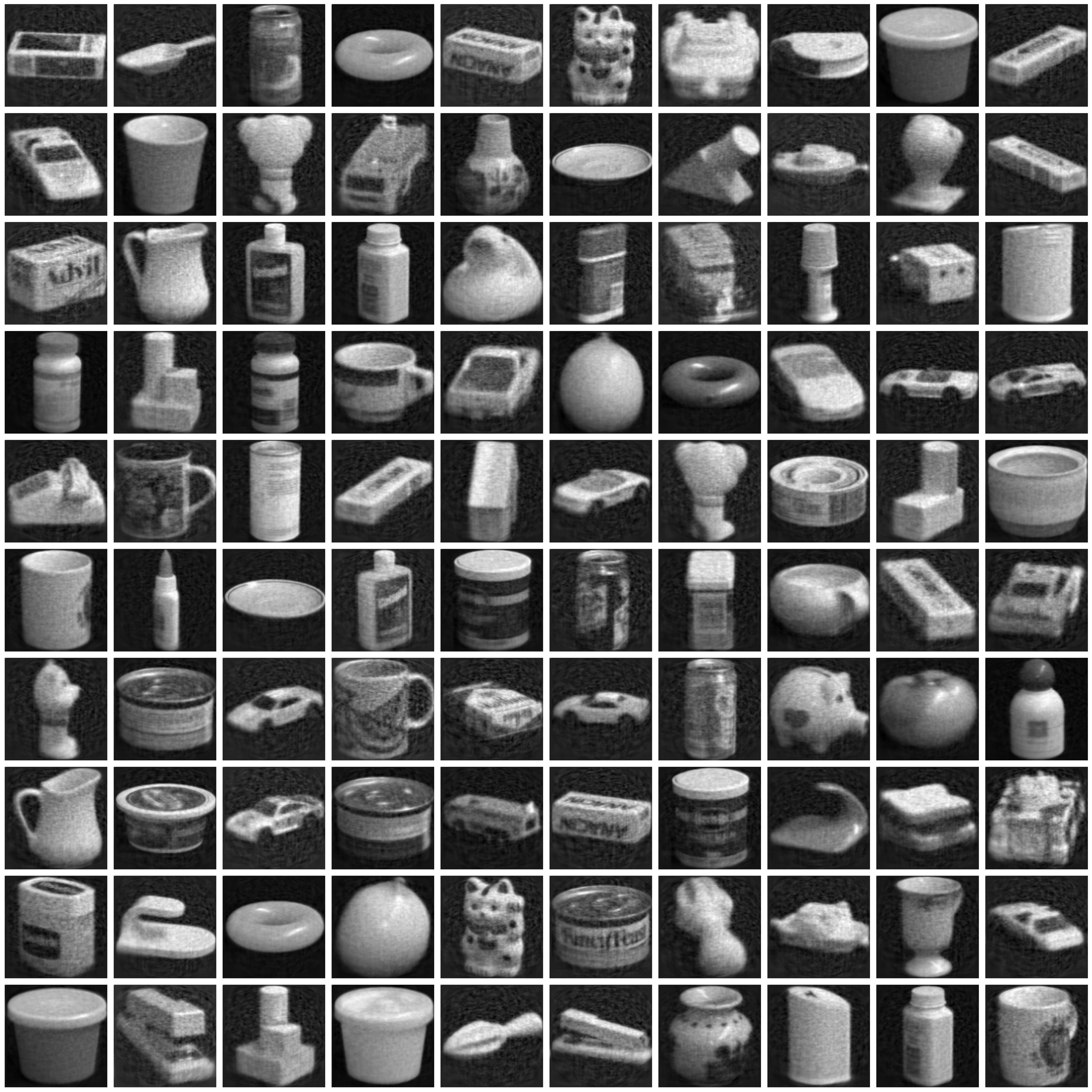}}\\
    \subfloat[Reconstruction using Kernel KSVD]{\includegraphics[height=5cm,width=9.3cm]{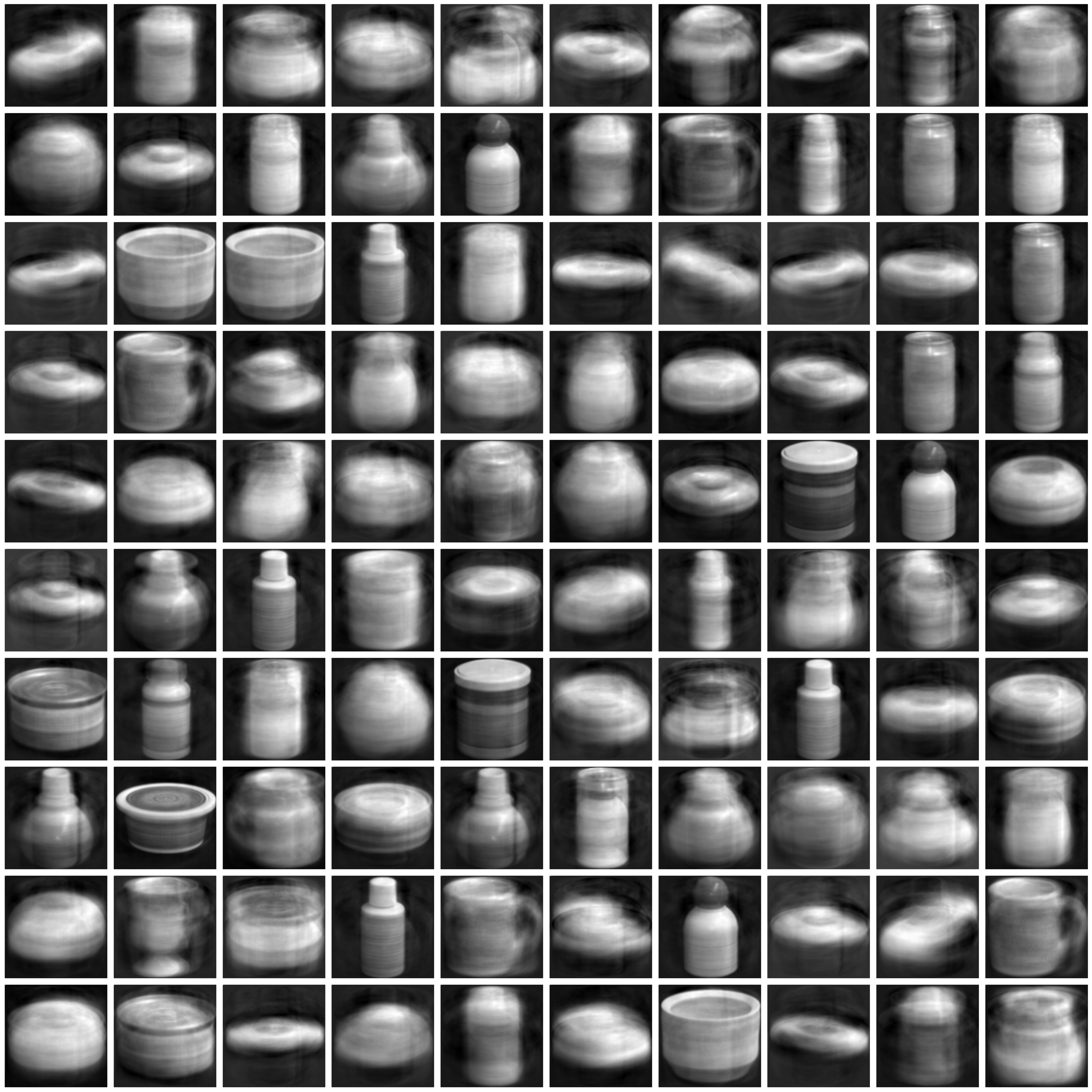}}
    \caption{COIL100 dataset: Comparison of reconstructions using RVFLDL and Kernel KSVD.}
    \label{fig:coil100_reconstr}
\end{figure*}

Columbia Object Image Library, COIL100 consists of images of  100 objects rotated through 360 degrees. Each object has 72 images of size $128\times 128$ with 5-degree pose variation \cite{COIL100nonlinearitydemonstrate}. From each class, around 62 images are used for training and 10 images for validation.
Figure \ref{fig:coil100_reconstr} shows the reconstructed COIL100 dataset images using RVFLDL with a Structural Similarity Index (SSI) close to 1 for each image in the dataset.
\subsubsection{Fashion-MNIST \cite{fashion-mnist}} 
\begin{figure*}
    \centering
    \includegraphics[height=5cm,width=10cm]{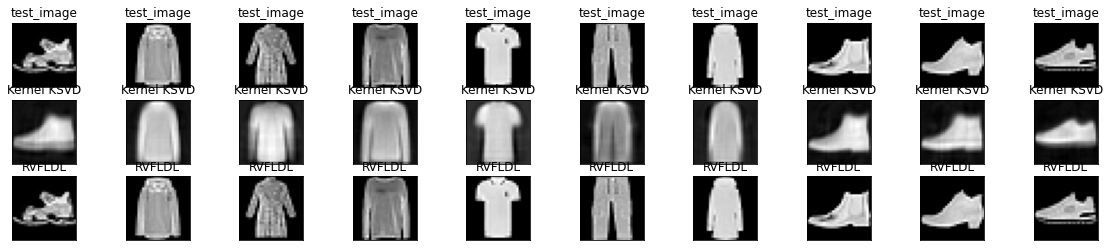}
    \caption{Fashion-MNIST test images in first column. Image reconstruction using Kernel KSVD (2nd row), RVFLDL (3rd row) with SSIs 1.}
    \label{fig:fashionmnist_reconstr}
\end{figure*}

The fashion-MNIST dataset from the database of Zalando's fashion clothing images has a training set containing $60,000$ images and a test set containing $10,000$ greyscale images of size $28\times 28$. The reconstructions of these images using RVFLDL are presented in Fig. \ref{fig:fashionmnist_reconstr}, and the structural similarity indices are equal to 1. 

\section{Discussion}\label{sec:rvflDiscuss}
 In RVFLDL, the dictionary size $K$ has to be much larger than the original dimensionality to attain overcompleteness and is increased after enhancing the patterns with their nonlinear transformations. However, this bottleneck is overcome through HS-prior as irrelevant features of the patterns are relegated to zero. Thus, assuming HS-prior over the coefficients makes the computation of the output matrix easier mitigating the computational complexity. RVFLDL with HS-prior reconstructs images with a high SSIM index. Assigning different levels of sparsity for global-local shrinkages enhances the classification performance of RVFLDL relative to other supervised nonlinear DL methods. The direct random connections between input and output layers act as a way of regularization. However, the global shrinkage parameter $\tau$ has to be decided empirically for better performance. 
Choosing the network parameters randomly takes care of the uncertainties in datasets. The mean of the $k$ classifiers and the mean of $k$ dictionaries obtained from k-fold cross-validation give a stable solution.

\section{Conclusions}\label{sec:rvflconclusion}
This paper presents a novel approach to learning nonlinear dictionaries using an SVD-free lightweight randomized RVFL to represent and classify nonlinear multivariate data efficiently and correctly. 
The dictionary is learned as a mapping from sparse coefficients to the dense input features. The nonlinear transformations of sparse coefficients induce nonlinearity in the dictionary atoms. Regularized Horseshoe prior assumed over the coefficients selects only relevant features, thus reducing the model complexity. Both the dictionary and the classifier matrix (for classification) are obtained as a single-step analytical solution to RVFL-net. This considerably reduces the computational overhead involved in other dictionary learning approaches which require computing SVD in each iteration. Unlike neural network-based nonlinear dictionary learning methods which require abundant data for training, our experiments show that lightweight RVFLDL performs well with fewer samples. Thus higher-order dependencies between the coefficients and the dictionary atoms are realized with limited computational resources. The experiments here demonstrate the effectiveness and the scalability of the method to high-dimensional datasets from a large number of classes with intra-class variance and inter-class similarities. The results imply the applicability of the method to different types and sizes of datasets. The complexity of the algorithm is lower and the classification accuracies are better than those of the existing methods.  In future works, the model could be applied to classify datasets with missing labels.

\bibliographystyle{elsarticle-harv}
\bibliography{rvfl}
\end{document}